\documentclass[sigconf, nonacm]{acmart}

\usepackage{float}
\usepackage{makecell}
\usepackage{subcaption}

\AtBeginDocument{%
  }

\setcopyright{acmlicensed}
\copyrightyear{2026}
\acmYear{2026}
\acmDOI{XXXXXXX.XXXXXXX}
\acmConference[Conference acronym ']{XXX}{XXXX}{XXXX}
\acmISBN{XXXXX}

\begin{document}

\title{From Prediction to Incrementality: Causal Optimization for Large-Scale Targeting and Recommendation}

\author{Changshuai Wei}
\authornote{These authors contributed equally}
\authornote{Corresponding Author}
\email{chawei@linkedin.com}
\affiliation{%
  \institution{LinkedIn}
  \city{Seattle}
  \country{USA}
}

\author{John Bencina}
\authornotemark[1]
\email{jbencina@linkedin.com}
\affiliation{%
  \institution{LinkedIn}
  \city{Sunnyvale}
  \country{USA}
}

\author{Phuc Nguyen}
\authornotemark[1]
\email{honnguyen@linkedin.com}
\affiliation{%
  \institution{LinkedIn}
  \city{Sunnyvale}
  \country{USA}
}

\author{Andre Assuncao Silva T Ribeiro}
\authornotemark[1]
\email{aribeiro@linkedin.com}
\affiliation{%
  \institution{LinkedIn}
  \city{Sunnyvale}
  \country{USA}
}

\author{Benjamin Zelditch}
\email{bzelditch@linkedin.com}
\affiliation{%
  \institution{LinkedIn}
  \city{Sunnyvale}
  \country{USA}
}

\begin{abstract}
Large-scale targeting and recommendation systems are typically built around predictive scores fed into heuristic or local allocation. When the business goal is incremental impact, as in marketing campaigns, incentives, and notifications, this paradigm systematically misallocates resources toward users who would have acted anyway. We present a decision-centric framework that instead optimizes causal effects under global constraints, aligning three components under a single objective: a causal neural network with a Transformer backbone for individual treatment-effect estimation, a Bayesian neural-bandit layer for uncertainty-aware exploration, and a dual-based large-scale linear-programming layer for constrained allocation. The framework also supports sequential context and multi-outcome, attribute-conditioned scoring through a Transformer encoder and outcome embeddings. We evaluate it with offline simulations on a public bandit dataset, targeted architectural ablations, and an online A/B test on LinkedIn Feed marketing traffic. We also distill production lessons on causal training-data construction and cost and delivery control, which were critical to successful deployment. The end-to-end treatment policy delivered a statistically significant $+7.20\%$ lift in the primary long-term-value metric, demonstrating the feasibility of production-scale causal optimization under business constraints.
\end{abstract}





\maketitle
\sloppy
% % Tighten spacing to fit page limit
% \setlength{\textfloatsep}{6pt plus 2pt minus 2pt}
% \setlength{\intextsep}{6pt plus 2pt minus 2pt}
% \setlength{\floatsep}{6pt plus 2pt minus 2pt}
% \setlength{\abovecaptionskip}{4pt}
% \setlength{\belowcaptionskip}{0pt}
% \setlength{\abovedisplayskip}{6pt plus 2pt minus 2pt}
% \setlength{\belowdisplayskip}{6pt plus 2pt minus 2pt}
% \setlength{\abovedisplayshortskip}{2pt plus 1pt}
% \setlength{\belowdisplayshortskip}{2pt plus 1pt}

\section{Introduction}
Large-scale targeting and recommendation systems are a core component of modern online platforms, supporting applications such as advertising, marketing outreach, notifications, and content recommendation. In practice, these systems are typically built by training predictive models to estimate user response probabilities (such as click-through or conversion likelihood) and then ranking or selecting items based on these predictions. This paradigm has proven effective for optimizing engagement metrics at scale and has become the dominant approach in both industrial deployments and academic research.

However, predictive modeling alone is fundamentally misaligned with the objectives of many real-world targeting problems. In marketing and incentive-driven settings, the goal is not to predict outcomes under historical policies, but to estimate the \emph{incremental impact} of an intervention relative to a counterfactual baseline. Observed user responses are confounded by prior targeting decisions, exposure mechanisms, and user self-selection, implying that high predicted response does not necessarily correspond to high causal value. Prior work has demonstrated that learning and evaluation based on biased recommendation logs can systematically misestimate the true effect of interventions when deployed for decision-making \cite{schnabel2016recommendations,wang2018deconfounded}.

This challenge has motivated increasing interest in \emph{causal machine learning} for recommender and targeting systems. A line of work reframes recommendations as treatments and user interactions as potential outcomes, enabling principled correction for exposure bias and confounding \cite{schnabel2016recommendations,bonner2018causal}. More broadly, causal perspectives on recommendation have been surveyed extensively, highlighting the limitations of purely predictive approaches and emphasizing the role of causal inference in reliable decision-making \cite{gao2024causalrecsys}.

In marketing and targeting applications, uplift modeling and treatment effect estimation explicitly target incrementality by estimating individual-level causal effects rather than response probabilities. Foundational work has studied the theoretical properties of individual treatment effect estimation and proposed representation-learning-based approaches to mitigate confounding bias \cite{shalit2017estimating}. More recent neural architectures, such as DragonNet, jointly model potential outcomes and treatment assignment, demonstrating improved stability and accuracy in treatment effect estimation \cite{shi2019adapting}.

Complementary to causal modeling, a rich literature studies counterfactual evaluation and learning from logged bandit feedback. Off-policy evaluation methods enable unbiased estimation of policy value without exhaustive online experimentation and have been extended to large action spaces and slate recommendation settings \cite{swaminathan2017offpolicy,mcinerney2020counterfactual}. While these approaches provide principled tools for evaluation and exploration, they typically focus on estimating or comparing policies rather than optimizing decisions for incremental outcomes.

At the same time, real-world targeting and recommendation systems must operate under complex global constraints, including budgets, capacity limits, frequency caps, and coverage or fairness requirements. To address these challenges, many industrial platforms rely on large-scale constrained optimization, often formulated as linear or mixed-integer programs, to translate model outputs into coordinated decisions. Theoretical foundations for robust and constrained optimization are well established \cite{bertsimas2011theory}, and industrial systems have demonstrated the effectiveness of combining learned utility models with constraint-aware optimization to improve system-level objectives \cite{agarwal2015constrained,makhijani2019lore,wei2024noah}.

Despite their success, existing optimization-based systems almost universally rely on \emph{predictive} scores or locally estimated uplift values as inputs. When incrementality matters, optimizing constrained decisions using non-causal or locally optimal signals can systematically misallocate resources toward users who would have acted regardless of intervention, thereby reducing true return on investment. Conversely, causal effect estimates alone are insufficient without a principled optimization layer to enforce constraints and optimize system-level objectives.

We present a \emph{decision-centric} formulation in which incrementality and constraints are addressed jointly: the system optimizes causal effects under global constraints, replacing the more common pattern of optimizing predictive or proxy signals under heuristic allocation. This reframes targeting as choosing users with high incremental impact and as globally coordinated allocation, rather than as ranking high-response users. The three components fit together naturally under this objective: causal modeling defines what is optimized, uncertainty-aware exploration shapes the data on which the causal estimates are learned, and constrained optimization translates those estimates into feasible decisions under shared resources. We instantiate the framework with a Transformer-augmented DragonNet causal head, a Bayesian neural-bandit layer, and a large-scale LP layer. While our primary motivation arises from marketing, the formulation is not domain-specific and extends to other settings that allocate limited intervention capacity under constraints.

Our contributions are: (i) a decision-centric formulation that aligns causal estimation, exploration, and constrained allocation under a single objective; (ii) two DragonNet architectural extensions, a Transformer encoder for marketing-touchpoint sequences and a shared outcome-embedding head, for multi-outcome and attribute-conditioned incremental scoring; and (iii) a productionized pipeline combining the causal head with neural-bandit exploration and a dual-based large-scale LP solver, evaluated on a public bandit dataset and as an end-to-end policy in a large online A/B test on LinkedIn Feed marketing traffic that delivered a $+7.20\%$ lift in the primary KPI ($p=0.041$).

\subsection{Related Work}
\label{sec:related_work}

Recent work has begun to incorporate treatment effect estimation and uplift modeling directly into recommender systems, moving beyond purely predictive objectives. Chen et al.~\cite{chen2024treatment} propose using individual treatment effect estimation to guide user interest exploration, demonstrating that causal signals can improve exploration efficiency compared to prediction-based uncertainty. Meng et al.~\cite{meng2025enhancing} introduce a coarse-to-fine dynamic uplift modeling framework for real-time video recommendation, showing that explicit uplift estimation can improve ranking quality at scale. Sun et al.~\cite{sun2024incentive} study incentive recommendation under budget constraints and propose an end-to-end uplift-based framework to improve cost-effectiveness.

While these approaches demonstrate the practical value of causal and uplift modeling in recommender systems, they primarily focus on \emph{local decision-making}, such as improving ranking quality or applying uplift-based heuristics under relatively simple constraints. Exploration is typically handled implicitly or through heuristic mechanisms, and optimization is performed at the level of individual users or items rather than as a coordinated system-level problem.

In contrast, our work integrates \emph{causal effect estimation, principled exploration, and constrained optimization} into a unified decision framework. We combine DragonNet-style causal modeling with Transformer-based sequential representations and a neural bandit layer, enabling uncertainty-aware exploration guided by estimated incremental effects. On the optimization side, we formulate targeting decisions as a large-scale constrained optimization problem and solve it using a dual-based approach that scales to extreme problem sizes. This allows us to coordinate decisions across users, items, and campaigns while enforcing complex global constraints, going beyond ranking-based or locally optimal uplift methods.

\section{Incremental Optimization Framework}

\subsection{Incremental Modeling}

\begin{figure*}[t]
\centering
\includegraphics[width=\textwidth]{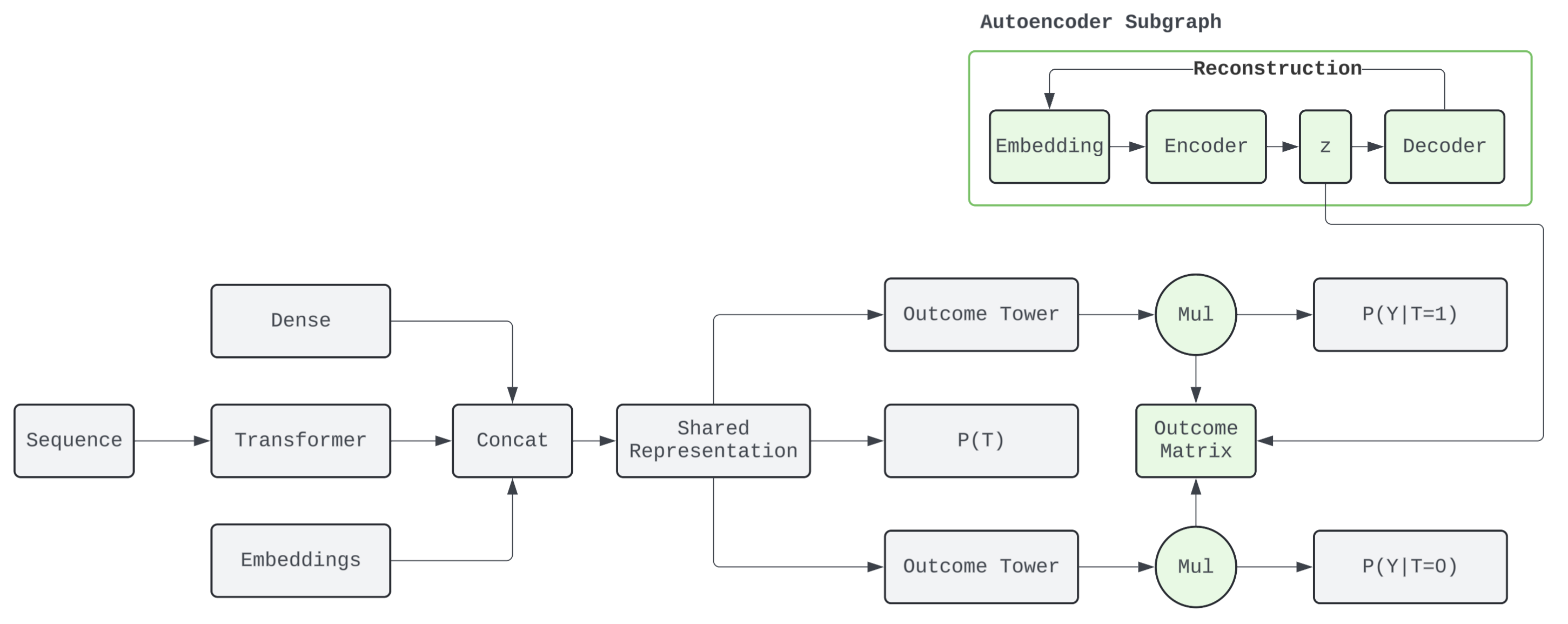}
\caption{Incremental model architecture overview. The model extends DragonNet with a transformer for temporal modeling and outcome embeddings for multi-product prediction. The core model utilizes member-level embeddings, dense features, and the marketing interaction sequence. The auxiliary autoencoder model utilizes product-level embeddings corresponding to each modeled outcome.}
\label{fig:model-architecture}
\end{figure*}

Incremental optimization requires estimating user-level treatment effects that quantify the expected lift from a targeting or recommendation action. Under the potential outcomes framework, let $Y(1)$ and $Y(0)$ denote the potential outcomes under treatment ($T=1$) and control ($T=0$), and let $X \in \mathcal{X}$ denote observed covariates. The Conditional Average Treatment Effect (CATE), or Individual Treatment Effect (ITE) at the covariate level, is

\begin{equation}
\tau(X) \;=\; \mathbb{E}[Y(1) - Y(0) \mid X] \;=\; \mu_1(X) - \mu_0(X),
\end{equation}

where $\mu_t(X) = \mathbb{E}[Y(t) \mid X]$. Two standard assumptions identify $\tau(X)$ from observational data: \emph{unconfoundedness}, $\{Y(0), Y(1)\} \perp\!\!\!\perp T \mid X$, and \emph{overlap}, $0 < e(X) < 1$ for all $X$, where $e(X) = P(T=1\mid X)$ is the propensity score. Under these, $\mu_t(X) = \mathbb{E}[Y \mid X, T=t]$ and $\tau(X)$ becomes point-identified from the observed data distribution.

Various estimators for $\tau(X)$ have been proposed, including S-Learner, T-Learner, X-Learner, and Double Machine Learning (DML) \citep{kunzel2019meta, chernozhukov2018double}. These approaches typically require multi-stage modeling pipelines, which introduce operational complexity at our scale of hundreds of millions of members. We instead adopt a modified version of DragonNet \citep{shi2019adapting}, shown in Figure~\ref{fig:model-architecture}, which jointly estimates $\mu_1(X)$, $\mu_0(X)$, and $e(X)$ in a single forward pass.

DragonNet extends TARNET \citep{shalit2017estimating} by adding a propensity head to a shared-representation architecture: a shared encoder $\Phi$ feeds two outcome heads estimating $\mu_1$ and $\mu_0$ (with observed-treatment gradient routing), and a third head estimating $e(X)$. The motivation is propensity sufficiency: under unconfoundedness, the propensity score is a balancing score, so any representation $\Phi$ that preserves $e(X)$ retains all information needed for CATE estimation. Outcome-only training tends to discard propensity signal in pursuit of marginal-likelihood fit, biasing $\hat{\tau}=\hat{\mu}_1-\hat{\mu}_0$ toward the dominant treatment arm. Jointly optimizing the propensity head therefore acts as an architectural regularizer on $\Phi$, and in practice yields markedly more stable CATE estimates~\citep{shi2019adapting}.

We extend DragonNet with three key enhancements for recommender systems. First, we feed sequential user-event features through a transformer layer (Section~\ref{sec:incrementalityTransformer}) before they enter $\Phi$, letting the encoder absorb temporal context such as recency and engagement cadence. Second, we extend the outcome heads to a multi-outcome configuration using learned product embeddings, so the model can score newly introduced products without adding a product-specific output head. Third, and more importantly, we couple this causal predictor with the neural bandit in Section~\ref{sec:exploration}, forming a \textit{DragonBandit} policy that explores over incremental action values rather than predicted responses.

\subsection{Neural Bandit Exploration}
\label{sec:exploration}

Reliance on logged data without any mechanism for exploration reinforces feedback loops, leading to ``rich-get-richer'' dynamics and suboptimal long-term performance~\cite{li2010contextual, su2024long, swaminathan2015batch, nguyen2026banditlp}. For causal modeling, exploration is additionally important to improve overlap and reduce dependence on the incumbent logging policy~\cite{schnabel2016recommendations, kasy2021adaptive}, supporting causal identification where the required assumptions hold~\cite{neal2020introduction}. At each round, DragonBandit draws posterior incremental scores rather than passing deterministic point estimates to the allocator in Section~\ref{sec:lp}. The multi-turn simulation examines this mechanism under feedback-loop bias.

Let $h_t$ denote the history and remaining constraint capacity at round $t$, $\mathcal{C}_t(h_t)$ the corresponding feasible set, and $x_t^*(\tilde{\tau}_t)$ its sampled-score solution. For nonzero exploration, the sampled logits have a nondegenerate Gaussian distribution; because the sigmoid heads are monotone, every open score ordering has positive probability. Therefore,
\begin{equation}
\pi_\alpha(i\mid u,h_t)=
\Pr\!\left[x^*_{u,i,t}(\tilde{\tau}_t)=1\mid h_t\right]>0
\label{eq:exploration_positivity}
\end{equation}
for every $(u,i)$ contained in some $x_t\in\mathcal{C}_t(h_t)$. Thus exploration guarantees positivity conditional on the feasible action.

\subsubsection{Neural Thompson sampling via Laplace approximation}
We approximate the posterior over network parameters by a Gaussian centered at the MAP solution $\hat{\theta}_{\mathrm{MAP}}$ via the linearized Laplace approximation (LLA)~\cite{su2024long,daxberger2021laplace,foong2019between,raha2024computationally}. The LLA computes the Gauss--Newton curvature of the log-likelihood at $\hat{\theta}_{\mathrm{MAP}}$ and propagates it through the network's local linearization, yielding a closed-form Gaussian on the logit $f_\theta(x^*)$ with mean $f_{\hat\theta}(x^*)$ and variance $\sigma_0^2 + g(x^*)^\top \Omega^{-1} g(x^*)$, where $g(x^*)$ is the Jacobian and $\Omega$ is the curvature matrix. Thompson sampling proceeds by drawing a logit sample and applying $\sigma(\cdot)$, so the same supervised checkpoint can be deployed with or without exploration and no retraining is required. Production scale is achieved through lightweight curvature approximations (low-rank, diagonal, or last-layer variants~\cite{nilsen2022epistemic,zhang2020neural,riquelme2018deep}); we use the last-layer variant for our pretrained DragonNet.

We also evaluated Bayes by Backprop (Appendix~\ref{app:bbb}) offline, but use LLA in production because it adds uncertainty sampling to the existing checkpoint with substantially lower training and serving overhead.

\subsection{Large-scale Allocation with Constraints}
\label{sec:lp}

The neural bandit improves decisions across rounds, but each round must still coordinate assignments under a common set of business constraints. Let $x_{u,i,t}$ indicate whether item $i\in\mathcal{I}$ is allocated to user $u\in\mathcal{U}$ at round $t=1,\ldots,T$. Using the sampled incremental objective $\tilde{\tau}_{u,i,t}^{\textrm{obj}}$ from DragonBandit, the horizon objective is
\begin{align}
    \mathop{\text{max}}_{x_{u,i,t}} &\sum_{t=1}^{T}\sum_{u,i}
    \tilde{\tau}_{u,i,t}^{\textrm{obj}}x_{u,i,t}, \label{eq:lp_primal}\\
    \text{s.t.} &\sum_{u,i}\tau_{u,i,t}^{\textrm{guardrail}_k}x_{u,i,t}
    \leq C_{\text{guardrail}_{k}}, \quad \forall\,t,k,\nonumber\\
                &\sum_i x_{u,i,t}\leq C_{\text{fcap}}, \quad \forall\,u,t,\nonumber\\
                &x_{u,i,t}\in\{0,1\}, \quad \forall\,u,i,t.\nonumber
\end{align}
Here $\tau_{u,i,t}^{y}=f_t^y(u,i,1)-f_t^y(u,i,0)$ is the round-specific sampled treatment effect on metric $y$. We use a sequential policy that solves one constrained optimization problem per round using the current posterior sample from DragonBandit. After each round, DragonBandit is updated using the observed feedback before producing the posterior sample for the next round. In the absence of shared allocation constraints, the resulting policy is equivalent to Thompson sampling: for each user, it selects the feasible action that maximizes the sampled incremental reward.

\subsubsection{Scalability via Dual Decomposition}

At production scale, each round contains tens of millions of users and hundreds of items, so its batch has $|\mathcal{U}|\times|\mathcal{I}|$ variables and is intractable for general-purpose solvers. Suppressing the round index, we relax $x_{u,i,t}$ to represent an action probability and solve the resulting large-scale problem using a smoothed dual-decomposition method \citep{basu2020eclipse}. The method adds a small ridge perturbation $\frac{\gamma}{2}\|x\|^2$ to the primal and dualizes the $K$ global constraints with multipliers $\lambda\geq 0$; by Danskin's theorem, the resulting dual $g_\gamma(\lambda)$ is differentiable with Lipschitz gradient $\nabla g_\gamma(\lambda) = A\,x_\gamma^*(\lambda) - b$, where the primal minimizer decomposes per user:
\begin{equation}
\label{eq:primal_step}
x_{\gamma,u}^*(\lambda) = \Pi_{\mathcal{C}_u}\!\left[-\tfrac{1}{\gamma}\bigl(A_u^T \lambda + c_u\bigr)\right],
\end{equation}
with $\Pi_{\mathcal{C}_u}$ an $O(|\mathcal{I}|\log|\mathcal{I}|)$ projection onto the per-user frequency-cap polytope. The solver then maximizes $g_\gamma$ over the $K$-dimensional dual with Nesterov-accelerated ascent, giving per-iteration cost linear in $|\mathcal{U}|\cdot|\mathcal{I}|$ versus $O((|\mathcal{U}||\mathcal{I}|)^{3.5})$ for interior-point methods.

The regularization $\gamma$ is picked as the largest value satisfying
$\frac{\gamma\, \hat{x}^T \hat{x}}{2\,|c^T \hat{x}|} < 10^{-3}$,
so the ridge perturbation contributes $<0.1\%$ of the objective and the perturbed optimum is practically indistinguishable from the true LP optimum; when $A$ is ill-conditioned across constraint scales, we apply Jacobi row preconditioning. In steady state, we warm-start the dual from the previous period's $\lambda^*$, which under stable input distributions (KS-tested) achieves over 99\% of the current optimum and also serves as an SLA fallback when the solver does not converge in time.

\section{Incrementality Transformer Architecture}
\label{sec:incrementalityTransformer}
\subsection{Temporal Features}

We construct a touchpoint sequence $S = (s_1, \ldots, s_L)$ over a lookback window, where each $s_i$ encodes a (channel, action) interaction such as (AD, IMPRESSION) or (EMAIL, OPEN); we also inject prior conversion events into the sequence. Each token is mapped through a learned embedding matrix $E_T \in \mathbb{R}^{V \times d}$, with $V$ interaction types plus a leading [CLS] token (BERT-style \citep{devlin2019bert}) that summarizes the sequence even when empty. We use the temporal encoding described below.

\paragraph{Temporal position.}
Member interactions occur on irregular schedules, so each token combines its interaction embedding with a day-since embedding $E_{DAYS}$, day-of-week embedding $E_{DOW}$, and the small-embedding-friendly sinusoidal positional encoding $PE_i$ of \citet{Foumani_2023}:
\begin{equation}
E_i = E_T(s_i) + E_{DAYS}(t_i) + E_{DOW}(\text{dow}_i) + PE_i.
\end{equation}
This represents both sequence order and elapsed time without expanding the interaction vocabulary. Appendix~\ref{app:position} gives the sinusoidal parameterization.

\paragraph{Multi-head attention.}
We apply standard multi-head self-attention~\citep{vaswani2017attention} with $Q = K = V = (E_1, \ldots, E_L)$, and pool by taking only the [CLS] output as the shared representation $E_S \in \mathbb{R}^d$, which is concatenated with member embeddings and dense features before $\Phi$.

\subsection{Multi-Outcome Extension}

Marketing campaigns generate multiple outcomes (products) per member, so we extend the single-outcome DragonNet to a multi-label setting $Y_k$, $k \in \{1, \ldots, K\}$. Because per-product conversion rates are highly imbalanced, we use inverse-frequency weights $c_k$ and combine the standard BCE loss (with sigmoid $\sigma$) into a per-head outcome loss
\begin{equation}
\mathcal{L}_{Y|T=t} = \tfrac{1}{K}\!\sum_{k} c_k\cdot\text{BCE}(y_k, \hat{y}_k),\quad \mathcal{L}_Y = \mathcal{L}_{Y|T=1} + \mathcal{L}_{Y|T=0},
\end{equation}
counting each head's loss only when the corresponding $T$ is observed (as in DragonNet/TARNET). Treatment prediction uses standard BCE on the propensity-head output $\hat{e}(X)$, $\mathcal{L}_T = \mathbb{E}[\text{BCE}(T, \hat{e}(X))]$.

Following \citet{shi2019adapting}, we add targeted regularization based on the efficient influence function (EIF) for the ATE under unconfoundedness~\citep{chernozhukov2018double}:
\begin{equation}
\phi_{\text{EIF}} = \hat{\tau}(X) + \tfrac{T}{\hat{e}(X)}(Y-\hat{\mu}_1)
- \tfrac{1-T}{1-\hat{e}(X)}(Y-\hat{\mu}_0) - \text{ATE}.
\end{equation}
The EIF is Neyman-orthogonal. We use its one-step correction as a training regularizer, not as a substitute for identification assumptions or as a stand-alone guarantee of unbiased CATE estimation:
\begin{equation}
\mathcal{L}_{\text{tarreg}} = \mathbb{E}\!\left[\bigl\| Y - (\hat{Y} + \epsilon\,\psi) \bigr\|^2\right],
\end{equation}
with $\psi = T/\hat{e}(X) - (1-T)/(1-\hat{e}(X))$ (the TMLE \emph{clever covariate} \citep{laan2006targeted}), learnable scalar $\epsilon$, and $\hat{Y}$ the prediction from the head matching observed $T$.

\subsection{Outcome Embeddings}

New-product launches create a cold-start problem: marketers want to promote an offering before the model has seen examples for it. Our use of ``CLIP-inspired'' refers narrowly to a normalized shared embedding space with temperature-scaled similarity; it does not use CLIP's large-scale contrastive pretraining. Instead of aligning images and text, we align a member's outcome state with an attribute-derived product representation, allowing the same member encoder to produce scores for products whose attributes can be embedded, including products unseen during training.

Concretely, each outcome tower (a two-dense-block MLP with linear projection) drops the final sigmoid and returns hidden logits $h_t(X), h_{nt}(X) \in \mathbb{R}^{d_o}$ representing the member's expected state under $T=1$ and $T=0$. A single shared outcome embedding matrix $E_O \in \mathbb{R}^{K \times d_o}$ encodes the $K$ products and is consumed by both towers, reflecting the fact that the product's intrinsic semantics are invariant to treatment.

\subsubsection{Outcome Autoencoder}

$E_O$ is the bottleneck of a lightweight autoencoder $E_O = \text{Encoder}(E^{\text{input}}_O)$ over outcome attribute representations. $E^{\text{input}}_O$ is constructed by mapping each product attribute through a learned embedding and concatenating; attributes can be either structured (one-hot product taxonomy, business-line, format) or unstructured (LLM-derived embeddings of product descriptions). The autoencoder is trained jointly with the main task using MSE reconstruction
\begin{equation}
\mathcal{L}_{\text{recon}} = \tfrac{1}{K}\!\sum_{k}\bigl\| E^{\text{input}}_{O,k} - \text{Decoder}(E_{O,k}) \bigr\|^2.
\end{equation}
This serves as semantic regularization: compared with a directly trained, task-specific $E_O$, the reconstruction objective encourages $E_O$ to retain the geometry of the input attribute space. The same setup enables large-scale embedding-space simulation, since pre-computed $h_t(X)$ and $h_{nt}(X)$ can be paired with arbitrary $E_O$ samples without re-running the full model.

\subsubsection{Outcome Matrix Layer}

The final outcome logits are L2-normalized dot products with learnable log-inverse-temperatures $\nu_t, \nu_{nt}$ (initialized to $\log 14.0$ following CLIP):
\begin{equation}
\hat{y}_t = \tfrac{h_t(X)}{\|h_t(X)\|} \cdot \tfrac{E_O^T}{\|E_O\|}\cdot\exp(\nu_t), \quad \hat{y}_{nt} = \tfrac{h_{nt}(X)}{\|h_{nt}(X)\|} \cdot \tfrac{E_O^T}{\|E_O\|}\cdot\exp(\nu_{nt}).
\end{equation}
Each row of $E_O$ is L2-normalized so that similarity depends purely on the angle between member-state and product representations, keeping logits on a consistent scale across products and treatment arms.

\subsection{Complete Loss Function}

The complete training objective sums the outcome, treatment, targeted-regularization, and reconstruction losses,
\begin{equation}
\mathcal{L} = \mathcal{L}_{Y|T=1} + \mathcal{L}_{Y|T=0} + \mathcal{L}_T + \mathcal{L}_{\text{tarreg}} + \lambda_{\text{recon}}\mathcal{L}_{\text{recon}},
\end{equation}
where $\lambda_{\text{recon}}$ weights the reconstruction loss; $\mathcal{L}_{\text{recon}}$ is included only when outcome attribute encodings are provided.

Although the complete research architecture contains several losses and optional modules, the serving path is modular rather than a jointly tuned monolith: outcome embeddings are used only for multi-product scoring, LLA is applied after supervised training to the last layer, and the LP consumes exported scores independently of model training. This separation lets each module be disabled or validated without retraining the rest of the decision pipeline, while the remaining tuning burden and feature sensitivity are limitations of the current shared-representation model.

\section{Offline Simulations}

In order to showcase the strengths of our proposed methodology, we perform an offline simulation study with a publicly available dataset. We focus on comparing the following approaches to targeting recommendations.

\begin{itemize}\setlength{\itemsep}{0pt}\setlength{\parskip}{0pt}
    \item Incremental Modeling + Constrained Optimization (ours): integrates uplift modeling with constrained optimization for system-level incremental targeting; uses dual decomposition~\citep{basu2020eclipse} at production scale and OR-Tools for simulation studies.
    \item Bandit Incremental Modeling + Constrained Optimization: extends our method with BNN-based exploration to address feedback-loop bias~\citep{nguyen2026banditlp}; same solver setup as above.
    \item Propensity Modeling + Constrained Optimization: optimizes predicted scores under constraints~\cite{agarwal2015constrained,makhijani2019lore} without modeling organic outcomes.
    \item Incremental Modeling + Ranking: ranks by estimated uplift~\cite{chen2020causalml,meng2025enhancing,sun2024incentive} but ignores global constraints.
    \item Propensity Modeling + Ranking: ranks by predicted response probability~\cite{li2010contextual,Covington2016YouTube,Cheng2016WideDeep,Ying2018PinSage}.
\end{itemize}

\subsection{Dataset}
We use the Open Bandit Dataset (OBD)~\cite{saito2020open}, a real-world logged bandit dataset. We utilize the random-policy subset, synthetically mapping the original 34 products that were recommended to $>400$K users to 5 distinct actions that are relevant to the production incrementality use case: recommendation to one of four business lines or no-recommendation. The no-recommendation action is a key difference between incremental and non-incremental targeting. We consider the logged reward within the OBD dataset as indicative of a conversion event to one of the four business lines. We augment that reward by coupling it with the average product price of the four business lines and apply per-business-line LP volume bounds, as shown in Table~\ref{table:bu_prices}.

\begin{table}[h]
    \centering
    \caption{Average product price and LP constraint bounds (as \% of audience) for each business line.}
    \label{table:bu_prices}
    \small
    \begin{tabular}{cccc}
        \hline
        \textbf{Business Line} & \textbf{Avg.\ Price} & \textbf{Min.\ Vol.} & \textbf{Max.\ Vol.} \\
        \hline
        A & \$5 & 5\% & 10\% \\
        B & \$10 & 5\% & 30\% \\
        C & \$10 & 5\% & 30\% \\
        D & \$200 & 30\% & 50\% \\
        \hline
    \end{tabular}
\end{table}

We also constructed a treatment variable that indicates whether a member was exposed to a marketing campaign showcasing one of the four business lines. This setup mirrors the production ecosystem and allows us to demonstrate the importance of incremental modeling. Finally, we assign a cost of \$0.1 to each recommendation/targeting action to capture operational and bidding costs.

\subsection{Setup}

We performed an 80/20 split of the full dataset into training and prediction sets. We trained the predictive models (incremental and propensity) on the training set, generating incremental/propensity scores for the prediction set. We applied constrained optimization or ranking to arrive at the final recommendations for the prediction set (one of the five actions described above), with the objective function constructed by coupling the incremental/propensity scores for conversion to each business line with the corresponding average product price.

We solve the offline simulation LP with Google OR-Tools, which is tractable at this scale (${\sim}400$K members, 5 actions) and convenient for reproduction; at full production scale we use the dual-decomposition method described in Section~\ref{sec:lp} instead. We apply the volume bounds from Table~\ref{table:bu_prices}. The ranking baseline assigns each member to the action with the highest predictive score, disregarding global business constraints.

We consider two simulation regimes: a \emph{single-turn} (static) evaluation and a \emph{multi-turn} (online) evaluation. For the single-turn setting, we construct the training log from a random logging policy in which all actions are approximately equally represented, and we evaluate each method once on the full prediction set using fixed model scores. For the multi-turn setting, we intentionally bias the training log by severely under-sampling a subset of actions, so that their estimated uplifts become high-variance and can even exhibit sign errors relative to the OBD ground-truth uplifts. We then simulate deployment over $T=200$ rounds using the prediction set as the environment: at each round, each method selects actions for the current batch, observes the realized rewards from its own recommendations, updates its training data accordingly, and incrementally updates the model before the next round.

\subsection{Results}

\subsubsection{Single-turn evaluation}

We evaluated the different methods by calculating the corresponding average rewards with Doubly Robust Policy Evaluation. As shown in Table~\ref{table:rewards}, incremental scores lead to higher rewards compared to propensity scores. We also provide the corresponding send volumes in Table~\ref{table:volumes}, where it can be seen that incremental targeting leads to a higher percentage of no-recommendations as the engine is able to identify members likely to convert organically.

\begin{table}[h]
    \centering
    \caption{Average Rewards, Costs, and Net Returns}
    \label{table:rewards}
    \label{table:costs}
    \label{table:returns}
    \small
    \setlength{\tabcolsep}{3pt}
    \begin{tabular}{ccccc}
        \hline
        \textbf{ML Model} & \textbf{Optim.} & \textbf{\thead{Avg \\ Reward}} & \textbf{\thead{Avg \\ Cost}} & \textbf{\thead{Avg Net \\ Return}}  \\
        \hline
        Incremental & Constr. Opt. & $\$0.55 \pm 0.14$ & $\$0.091 \pm 0.001$ & $\$0.46 \pm 0.13$ \\
        \hline
        Propensity & Constr. Opt. & $\$0.49 \pm 0.18$ & $\$0.1 \pm 0$ & $\$0.39 \pm 0.18$ \\
        \hline
        Incremental & Ranking & $\$0.46 \pm 0.15$ & $\$0.090 \pm 0.001$ & $\$0.37 \pm 0.15$ \\
        \hline
        Propensity & Ranking & $\$0.41 \pm 0.20$ & $\$0.1 \pm 0$ & $\$0.31 \pm 0.20$ \\
        \hline
    \end{tabular}
\end{table}

\begin{table}[h]
    \centering
    \caption{Send Volumes}
    \label{table:volumes}
    \small
    \begin{tabular}{cccccc}
        \hline
        \textbf{Method} & \textbf{\thead{Bus. \\ Line \\ A}} & \textbf{\thead{Bus. \\ Line \\ B}} & \textbf{\thead{Bus. \\ Line \\ C}} & \textbf{\thead{Bus. \\ Line \\ D}} & \textbf{\thead{No \\ Rec.}} \\
        \hline
        \makecell{Incremental + \\ Constr. Opt.} & $5\%$ & $30\%$ & $26\%$ & $30\%$ & $9\%$ \\
        \hline
        \makecell{Propensity + \\ Constr. Opt.} & $5\%$ & $18\%$ & $27\%$ & $50\%$ & $0\%$ \\
        \hline
        \makecell{Incremental + \\ Ranking} & $0\%$ & $46\%$ & $44\%$ & $0\%$ & $10\%$ \\
        \hline
        \makecell{Propensity + \\ Ranking} & $0\%$ & $0\%$ & $0\%$ & $100\%$ & $0\%$ \\
        \hline
    \end{tabular}
\end{table}

The send volumes also highlight the difference between Constrained Optimization and Ranking. As expected, Ranking provides solutions that deviate significantly from the desired business constraints.

We show the average costs in Table~\ref{table:costs}, where it can be seen that incremental modeling leads to reduced cost due to the reduced send volumes described above. The corresponding net returns are shown in Table~\ref{table:returns}, where Incremental Modeling coupled with Constrained Optimization provides the highest overall performance.

\subsubsection{Multi-turn evaluation}

\begin{figure}[h]
\centering
\begin{subfigure}[b]{0.49\columnwidth}
\centering
\includegraphics[width=\linewidth]{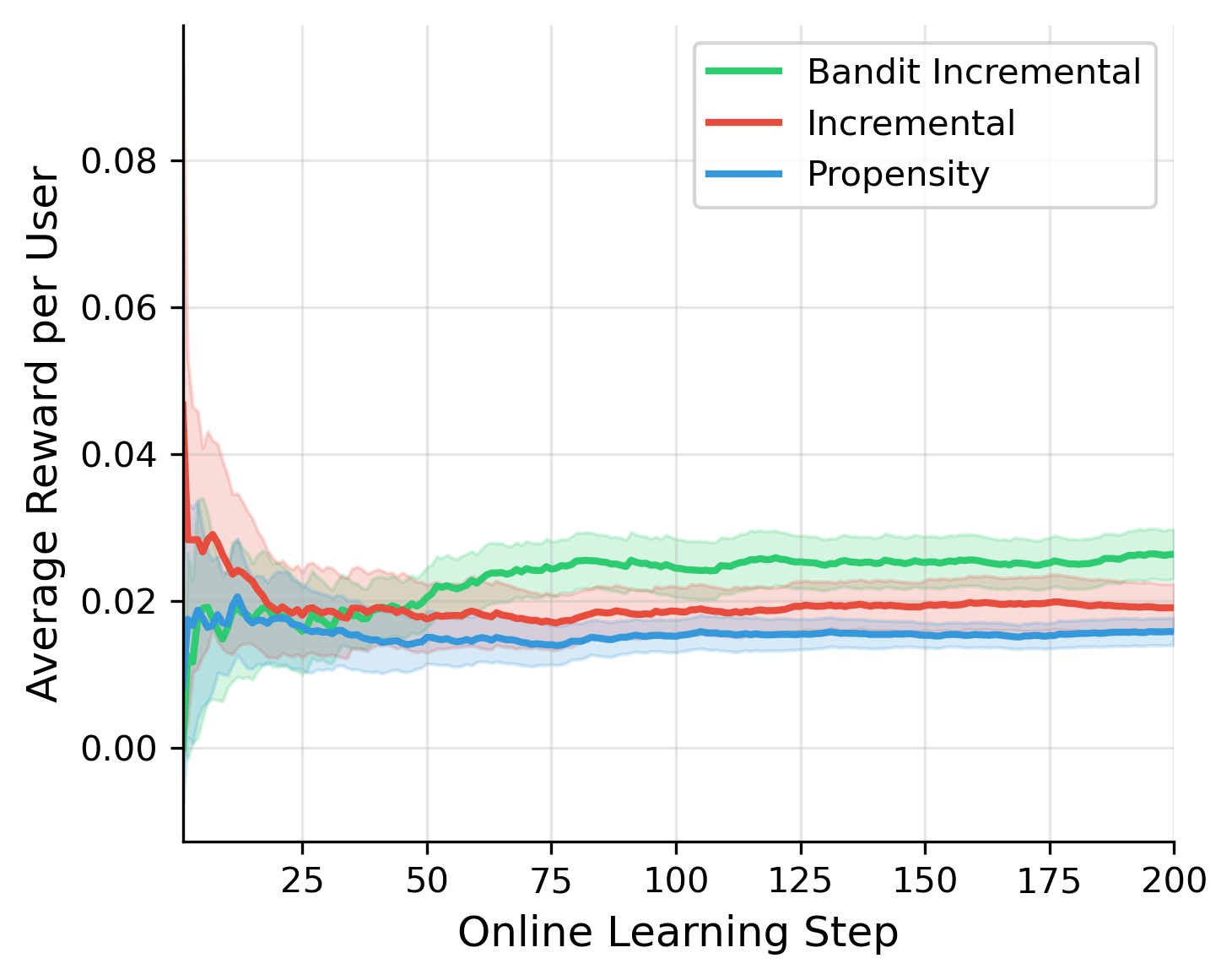}
\caption{Small bias setting.}
\label{fig:multi-turn-ee-small-bias}
\end{subfigure}
\hfill
\begin{subfigure}[b]{0.49\columnwidth}
\centering
\includegraphics[width=\linewidth]{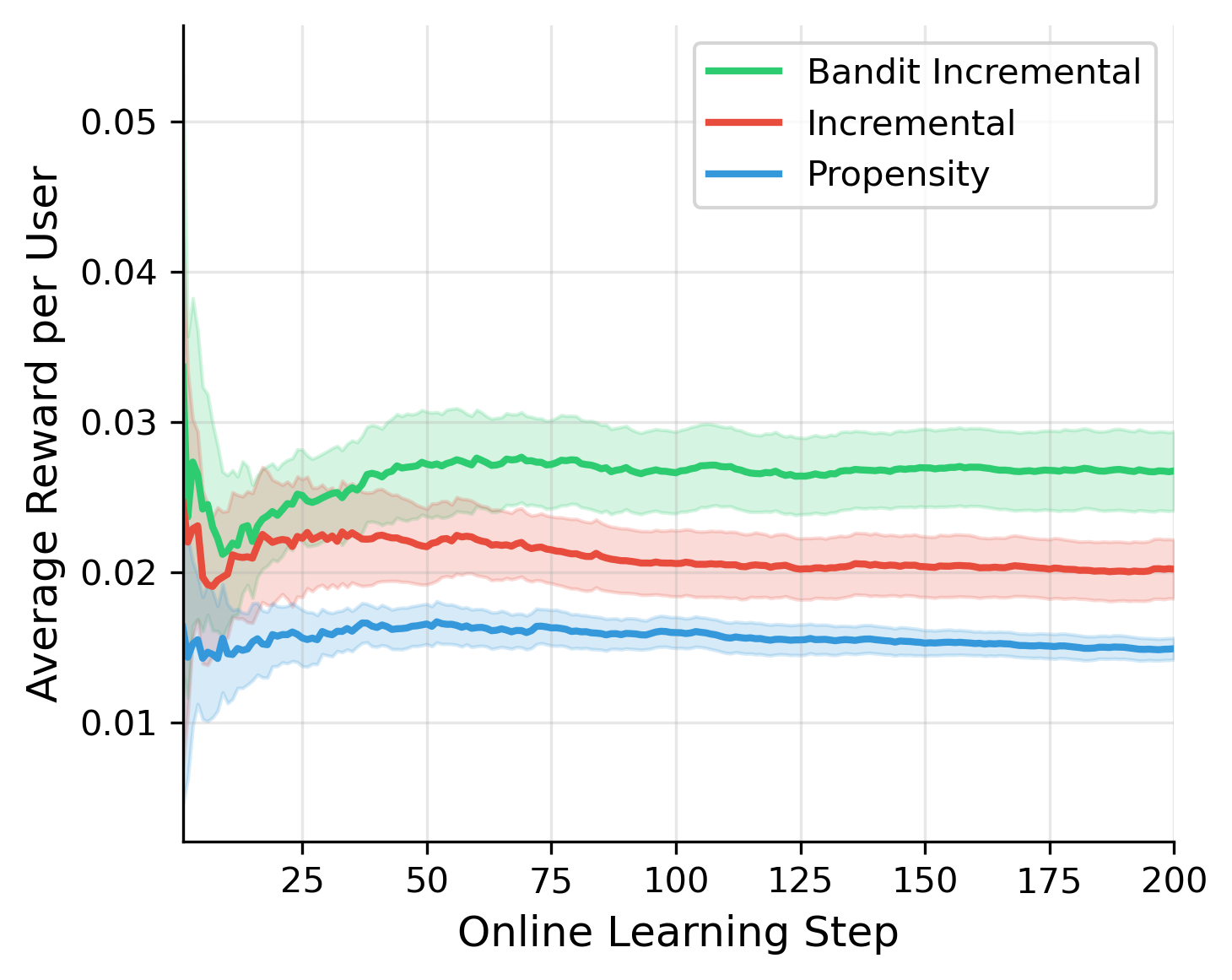}
\caption{Large bias setting.}
\label{fig:multi-turn-ee}
\end{subfigure}
\caption{Average cumulative reward and 95\% confidence intervals after 200 rounds of feedback. The confidence intervals are calculated from 30 simulation runs.}
\end{figure}

The previous single-turn evaluation shows that only approaches with Constrained Optimization can adequately satisfy the volume constraint bounds induced by business constraints, so we restrict attention to these approaches in the multi-turn evaluation. This multi-turn evaluation demonstrates the advantage of having an explicit mechanism to balance exploration and exploitation at the system level. The initial training-data bias we inject is designed to mirror common challenges in industry applications, such as cold-start, low representation for certain actions, and non-stationarity where the true reward function evolves over time (e.g., an initially low-performing action later becomes high-performing). Figure~\ref{fig:multi-turn-ee-small-bias} shows that the Bandit Incremental Model is able to overcome this bias and ultimately outperform the other approaches after learning from the outcomes of its own actions, even though it may underperform its greedy counterpart in the first few steps of online learning. This reflects the short-term cost of exploration in exchange for long-term gains: after roughly 50 model updates, the Bandit Incremental Model begins to outperform both greedy variants. Finally, we show that the benefit of exploration is more pronounced when the initial training-data bias is larger, as illustrated in Figure~\ref{fig:multi-turn-ee}.

\subsection{Ablation Study}

\begin{figure*}[t]
\centering
\begin{subfigure}[b]{0.32\textwidth}
\centering
\includegraphics[width=\linewidth]{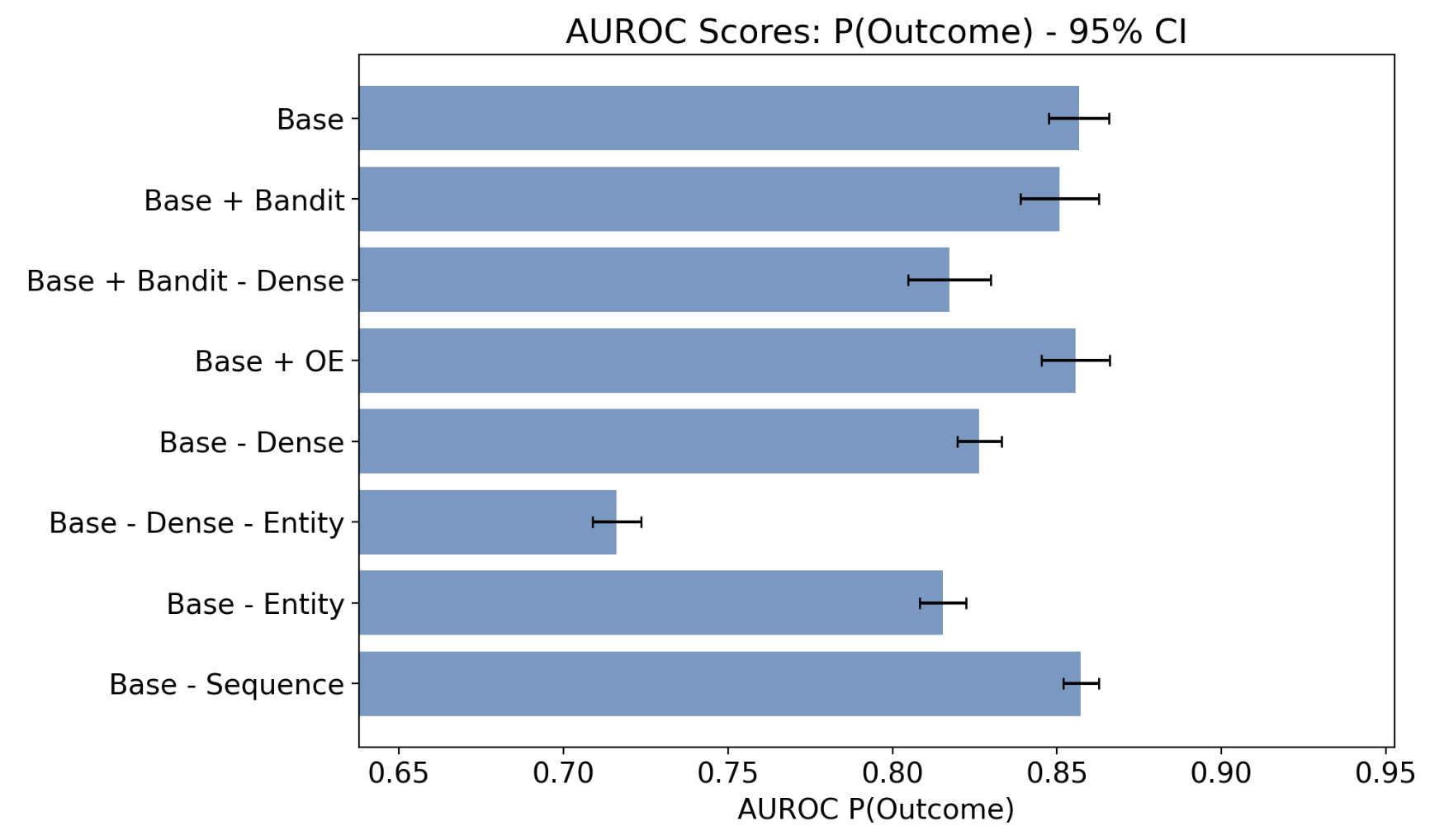}
\caption{Outcome prediction AUROC.}
\label{fig:ablation_outcome}
\end{subfigure}
\hfill
\begin{subfigure}[b]{0.32\textwidth}
\centering
\includegraphics[width=\linewidth]{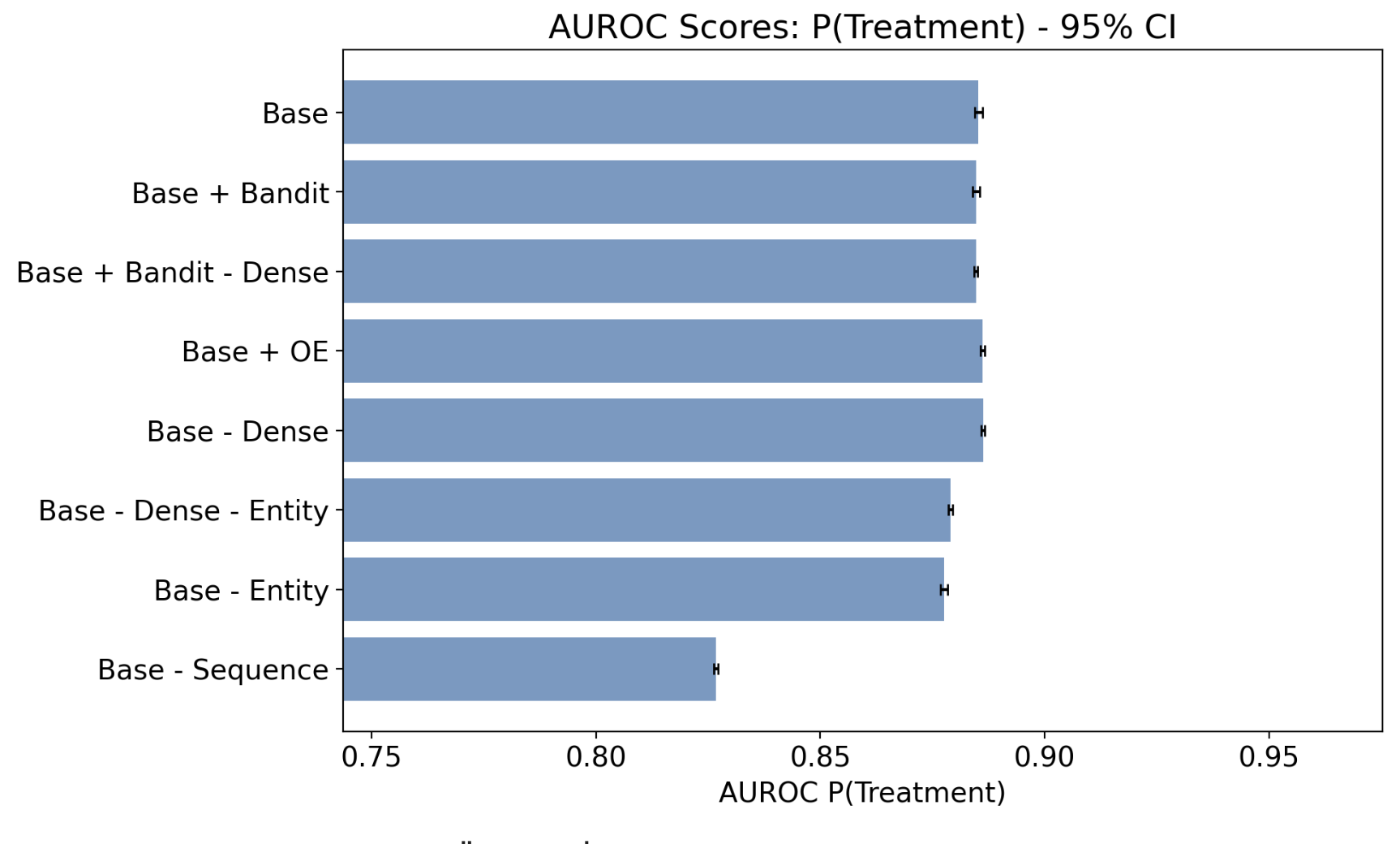}
\caption{Treatment prediction AUROC.}
\label{fig:ablation_treatment}
\end{subfigure}
\hfill
\begin{subfigure}[b]{0.32\textwidth}
\centering
\includegraphics[width=\linewidth]{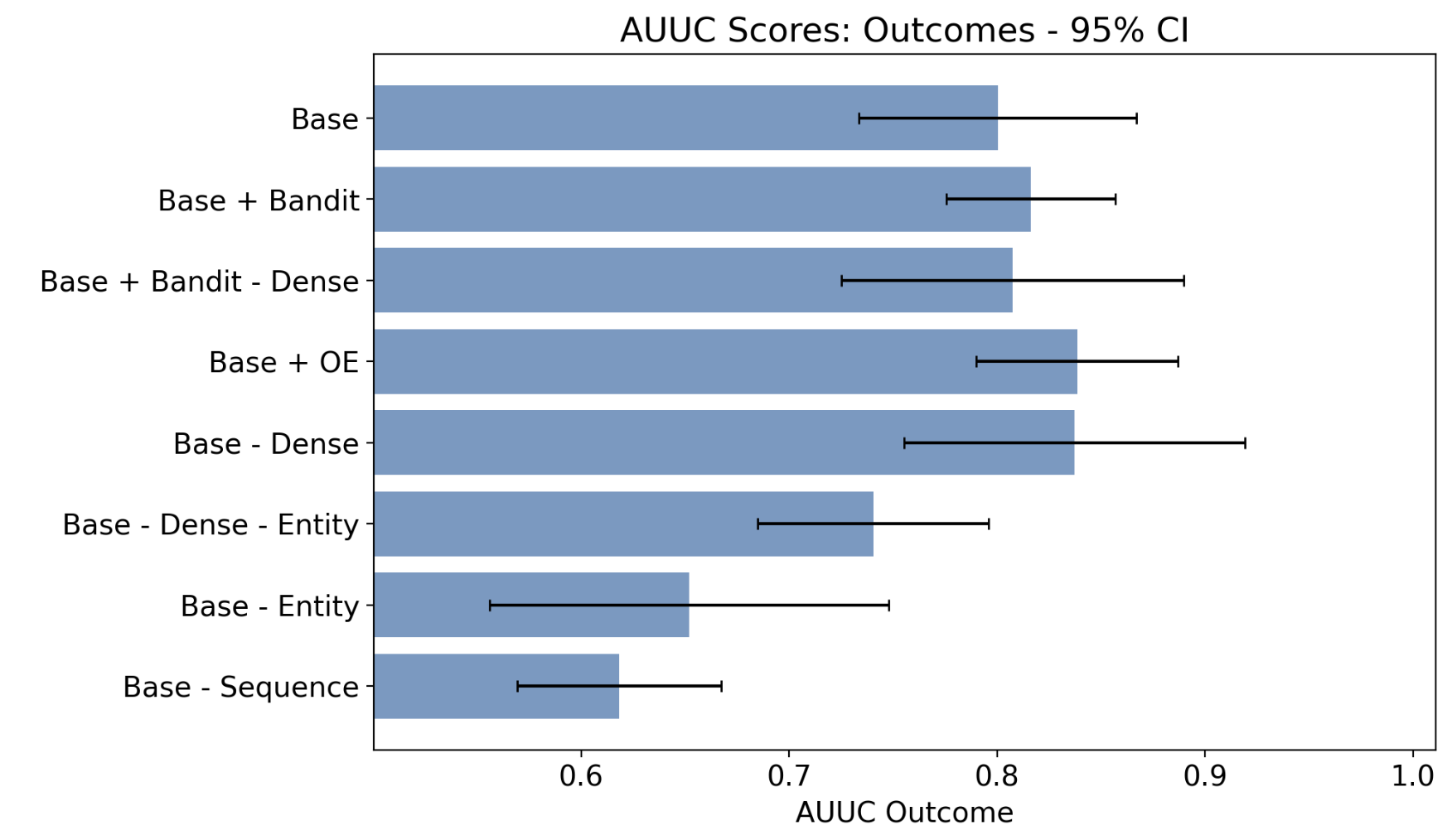}
\caption{Uplift AUUC.}
\label{fig:ablation_auuc}
\end{subfigure}
\caption{Ablation results. Error bars denote 95\% confidence intervals from 5 training runs.}
\end{figure*}

We tested 8 configurations on a fixed train/validation snapshot, each repeated 5 times with common hyperparameters. Figure~\ref{fig:ablation_outcome} summarizes outcome, treatment, and uplift performance. Adding outcome embeddings preserves outcome AUROC and produces the highest uplift AUUC point estimate. Removing dense features reduces outcome AUROC, while its uplift effect depends on the configuration: AUUC increases for the base model but decreases slightly when bandit exploration is enabled. This is consistent with the dense features being prognostic rather than treatment-effect modifiers. Appendix~\ref{app:ablation} provides the full methodology, results, and outcome-embedding diagnostic.

\section{Production Deployment and Experimentation}
\label{sec:deployment}
We now turn from modeling to \emph{production}. Deploying causal targeting at scale taught us that a handful of pragmatic enhancements beyond the core model and optimizer are what make the system work in practice. The most important are the construction of training data and the control of spend and delivery. These enhancements are largely invisible to offline simulations and ablation studies, which hold the data-generating process and delivery pipeline fixed, yet in our experience they are important for real-world success. We describe these enhancements first, followed by the serving architecture, agentic experimentation setup, and online results.

\subsection{Training-Data Construction for Causal Estimation}
\label{sec:trainingdata}

Production logs allow us to build sequences of marketing touches and conversions, but causal estimation requires careful construction of the context sequence $X$, treatment action $T$, and conversion label $Y$. The context must contain only information available before treatment, while treatment and outcome are assigned from subsequent non-overlapping windows.

Let $R$ denote the snapshot run date. For each member, production samples $D=R-(W_C+W_T+W_D)+U$, where $U\sim\operatorname{Uniform}\{0,\ldots,W_D-1\}$, $W_D=90$ days, $W_T=7$ days, and $W_C=30$ days. Thus $D\in[R-127,R-38]$. The model input contains marketing interactions and prior conversions in $[D-60,D)$; post-$D$ events are excluded. We set $T=1$ when at least one qualifying email send, on-platform impression, or video view occurs in $[D,D+7)$. We set $Y=1$ when the corresponding business-line or product-family conversion occurs in $[D+7,R]$, and $Y=0$ otherwise. This reserves at least 30 days of follow-up while earlier dates have longer outcome windows. Randomizing $D$ avoids a last-touch label, captures long-term action effects, and preserves variable-length histories.

\subsection{Cost and Delivery Control}
\label{sec:costcontrol}

Controlling how much the causal policy spends and delivers matters for two reasons: a standing \emph{business} requirement to pace committed budgets over the fiscal quarter, and an \emph{experimentation} requirement to compare arms at matched delivery. Both stem from the same underlying gap between an assignment and its realized delivery.

At a given round, with $t$ suppressed, the LP in~\eqref{eq:lp_primal} chooses assignments $x_{u,i}$, but an assignment is not a guaranteed delivery in an auction-mediated, activity-gated channel. A member may not return during the campaign window, or the send may lose the downstream auction. We therefore estimate the probability of delivery,
\begin{equation}
p_{u,i}=P(\text{delivered within window}\mid u,i),
\end{equation}
using a classifier with isotonic calibration per member-lifecycle segment. The estimate enters the LP through the expected-impression constraint $\sum_{u,i}p_{u,i}x_{u,i} \in [C^l_{\text{imp}}, C^u_{\text{imp}}]$ and the expected-cost constraint $\sum_{u,i}\hat c_{u,i}x_{u,i} \in [C^l_{\text{cost}}, C^u_{\text{cost}}]$. Their bounds are calibrated to the observed BAU delivery envelope. This prevents the optimizer from concentrating assignments on members who appear incremental but are difficult or expensive to reach.

Across multiple rounds, such as in quarterly budget pacing or cumulative spend matching in an experiment, realized delivery and cost may differ from their LP predictions. We therefore introduce a feedback controller that adjusts the round-level cost target. Let $B$ be the committed budget, $S_t$ the cumulative realized spend, and $q(t/H)$ the desired cumulative pacing curve over a horizon $H$, with $q(0)=0$ and $q(1)=1$. The outer loop computes
\begin{equation}
S_t^\star=Bq(t/H), \qquad e_t=S_t^\star-S_t .
\end{equation}
Let $D_t$ denote the realized spend during interval $t$, and let $A_t=\sum_{u,i}\hat c_{u,i,t}x_{u,i,t}$ denote the LP-predicted spend of the selected assignments. The inner loop estimates the spend-realization ratio as $r_t=D_t/A_t$ and smooths it using $\hat r_t=\alpha r_t+(1-\alpha)\hat r_{t-1}$, where $\alpha\in(0,1]$ weights the newest observation. It then updates the LP cost target:
\begin{align}
C_{\text{cost},t+1}
&=
\Pi_{[C_{\min},C_{\max}]}\left(
C_{\text{cost},t}
+\kappa\frac{e_t}{\max(\hat r_t,\epsilon)}
\right),
\end{align}
where $\kappa>0$ is the controller gain. The target $C_{\text{cost},t}$ can be converted into a two-sided LP constraint using a predefined tolerance around the target.

When realized spend falls behind the pacing curve, the controller raises the assignment cap in proportion to the gap and estimated deliverability; when spend runs ahead, it lowers the cap. The projection preserves operational bounds.

An early A/B test run \emph{without} the above controls showed the treatment arm under-delivering: the causal policy withholds and reallocates sends, lowering treatment impressions relative to control. A raw total-bookings comparison then penalizes the treatment for delivering \emph{less} rather than for choosing worse, confounding policy quality with delivery volume. Constraining treatment to the BAU impression and cost envelope equalizes delivery across arms and restores a clean, like-for-like read.

\subsection{Serving Architecture}
\label{sec:serving}
The system is served as a batch pipeline. DragonBandit scores $\mu_1$, $\mu_0$, and $e$ for the eligible population and samples incremental objectives via last-layer linearized Laplace Thompson sampling (Section~\ref{sec:exploration}); a dual-decomposition solver~\citep{basu2020eclipse} computes the constrained allocation~\eqref{eq:lp_primal} under global guardrails, warm-started from the previous period's dual $\lambda^\star$; and the feedback controller (Section~\ref{sec:costcontrol}) adjusts budget caps from realized deliverability before assignments are materialized to the campaign-management system. Because scoring and allocation are decoupled from serving, the same supervised checkpoint is deployed with or without exploration, and non-convergence solver falls back to the warm-started dual, which recovers over 99\% of the current optimum under stable input distributions within the SLA.

\subsection{Agentic Experimentation}
\label{sec:expdesign}
We evaluated the end-to-end framework with an eight-week online A/B test on LinkedIn Feed marketing traffic, against the deployed business-as-usual (BAU) targeting stack. BAU is a standard two-tier recommender: the retrieval tier pre-selects each campaign's audience using a propensity-based scoring model together with marketer-defined criteria, and the ranking tier is a second propensity-based model that estimates engagement probability and is followed by ranking heuristics for final assignment. The treatment arm keeps the marketer-defined criteria in retrieval and replaces both propensity layers with a single decision layer that scores members by predicted causal uplift across all campaigns and assigns them via the LP under global constraints. A meaningful policy difference is that the treatment arm can \emph{withhold} a send whenever the predicted incremental value is negative or no feasible positive-incremental option exists. Members were randomly assigned 50/50 to the two experiment arms.

Standing up a \emph{representative} BAU control and a matched treatment arm at scale requires configuring hundreds of live campaigns across products, segments, and budgets. Manual configuration is infeasible and, we found, subject to two subtle failure modes that bias measurement yet are invisible offline. First, when out-of-scope campaigns are suppressed via an exclusion segment defined by \emph{dynamic} criteria (company, locale, activity), members drift across arms in a way correlated with platform activity, contaminating the intent-to-treat contrast. Second, when arms are assembled from shared segment definitions, a control audience can silently inherit a \emph{treatment} send decision through a reused sub-segment. We therefore built an agentic campaign-setup tool that duplicates campaigns into hundreds of budget-split variants, \emph{freezes} each audience to a static snapshot for the experiment's duration, audits segment lineage so that no control segment depends on a treatment-derived decision, and materializes campaigns as drafts promoted by a single activation toggle with symmetric rollback. This provides identical scaffolding across arms, so the only systematic difference is the decision layer. It also improves operational safety. Budget parity is enforced by setting caps proportional to audience sizes and recalibrating as arm sizes drift.

\subsection{Results}
\label{sec:results-online}
Measured against long-term-value metrics over an eight-week period, the treatment arm achieved a statistically significant $+7.20\%$ lift ($p = 0.041$, 95\% CI: $[0.31\%, 14.09\%]$). The experiment evaluates the end-to-end policy as deployed. Its components satisfy coupled system requirements rather than representing independent product features: causal scoring defines the incremental objective, the exploration policy provides stochastic treatment variation that improves positivity (overlap), the LP enforces global business constraints, and the withhold rule prevents assignments with negative incremental value. Removing any of these components changes the operating requirements or the data-generating policy. The online result therefore measures the system-level impact of the complete production policy, while the offline studies examine individual mechanisms under controlled settings.

\appendix
\section{Detailed Ablation Study}
\label{app:ablation}

\subsection{Protocol}
We ran targeted ablation simulations on a static train/validation snapshot, holding hyperparameters (learning rate, epochs, batch size, hidden sizes) constant across configurations and varying only the indicated change. We tested 8 configurations, each repeated 5 times. The base model is the architecture from Section~\ref{sec:incrementalityTransformer} on top of the DragonNet causal head, without bandit sampling or outcome embeddings (OE); we then added bandit and OE separately, and additionally removed three input categories: dense features, entity embeddings, and sequential interactions. These are not exhaustive component ablations: in particular, they do not separately isolate the Transformer operator from its sequence input, targeted regularization, or every interaction among loss terms.

\subsection{Outcome and Treatment Metrics}
Outcome ROC-AUC is computed against the common validation dataset using the observed treatment for masking, $\hat{y}=T\hat{y}_t+(1-T)\hat{y}_{nt}$, and macro-averaged across labels. Treatment AUROC evaluates the propensity head against the observed treatment assignment.

\subsection{Uplift Results}
Following CausalML~\citep{chen2020causalml}, we rank observations by predicted uplift and compute
\begin{equation}
\text{CumLift}(k) =
\tfrac{\sum_{i=1}^{k}Y_iT_i}{\sum_{i=1}^{k}T_i+\epsilon}
-\tfrac{\sum_{i=1}^{k}Y_i(1-T_i)}{\sum_{i=1}^{k}(1-T_i)+\epsilon},
\end{equation}
with $\text{CumGain}(k)=k\,\text{CumLift}(k)$ and normalized area under the gain curve as AUUC. Implementations of AUUC and Qini differ~\citep{gutierrez2017causal,radcliffe2007using,devriendt2020learning}; we use the CausalML implementation.

Removing dense features is neutral or beneficial for uplift AUUC despite reducing outcome AUROC from 0.857 to 0.826 and minimally affecting treatment AUROC. A plausible explanation is that these features are prognostic and predict outcomes regardless of treatment, rather than serving as treatment-effect modifiers. They may therefore add variance when the shared DragonNet representation must serve both propensity and outcome losses. This remains a hypothesis rather than an automated feature-selection mechanism; architectures such as FlexTENet~\citep{curth2021inductive} could explicitly separate the two subspaces.

\subsection{Outcome-Embedding Diagnostic}
\begin{figure}[h]
\centering
\includegraphics[width=0.75\columnwidth]{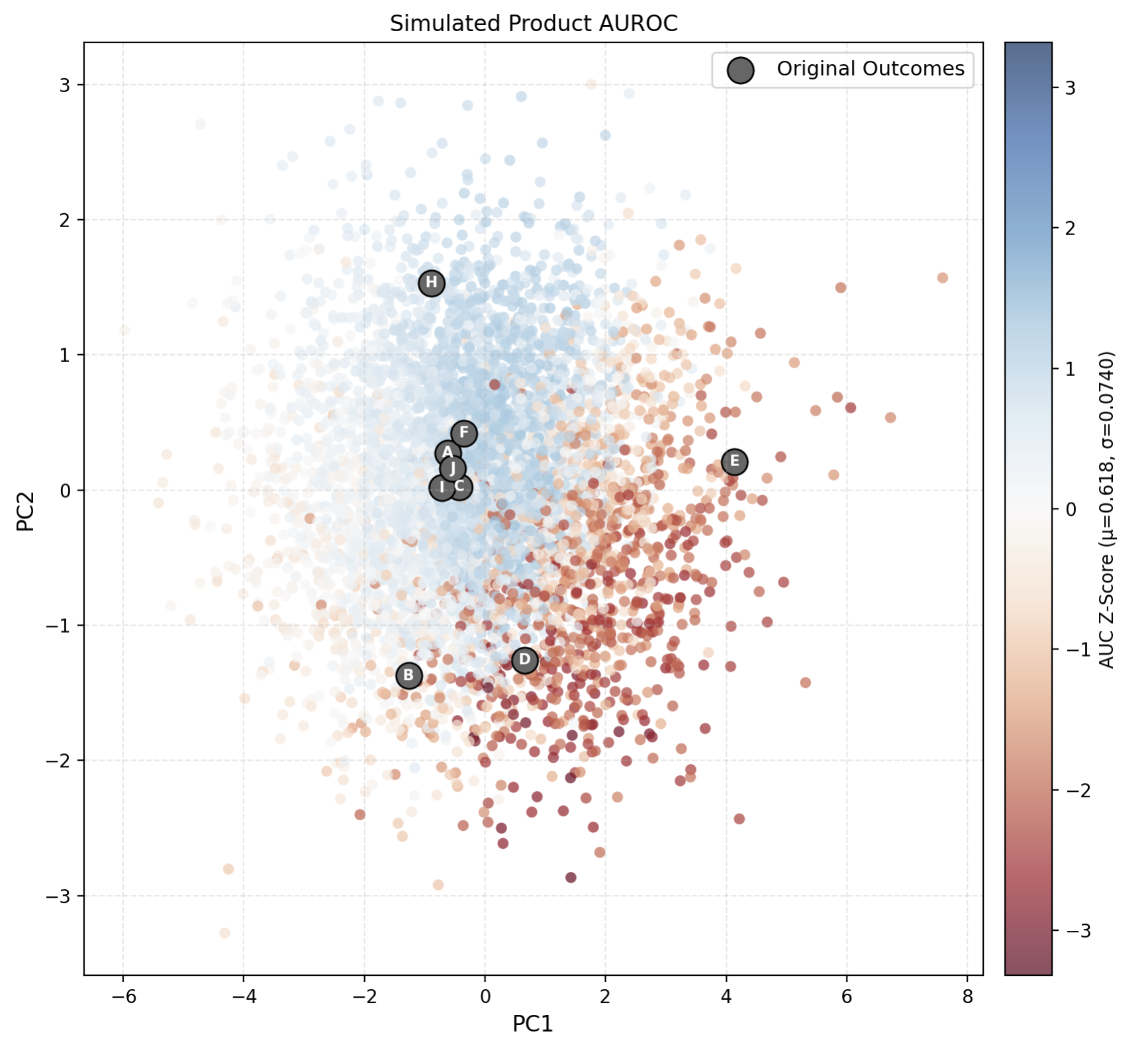}
\caption{Outcome-embedding PCA. Each sampled hypothetical product is colored by AUROC z-score against the held-out product.}
\label{fig:simulation-plot}
\end{figure}

We train on 9 of 10 outcomes, draw 5{,}000 hypothetical products from the learned embedding space, and score each against the held-out 10th product across the validation members. Figure~\ref{fig:simulation-plot} shows that samples near semantically similar trained products have higher AUROC against the held-out target, indicating that the geometry retains outcome-relevant structure. Because this diagnostic uses one held-out product and sampled embeddings rather than prospective real launches, it does not establish zero-shot production effectiveness.

\section{Additional Architectural Details}

\subsection{Bayes by Backprop}
\label{app:bbb}
Bayes by Backprop~\cite{blundell2015weight,kingma2015variational} learns a Gaussian variational posterior $q_\phi(\theta)$ by maximizing the evidence lower bound. Predictive probabilities are approximated by Monte Carlo sampling,
\begin{equation}
P(y^*=1\mid x^*) \approx \frac{1}{M}\sum_{m=1}^{M}\sigma(f_{\theta^{(m)}}(x^*)),
\qquad \theta^{(m)}\sim q_\phi.
\end{equation}
Compared with LLA, it captures non-local posterior uncertainty but adds stochastic training and repeated inference-time sampling. For a minibatch variant, an additional $\beta\mathcal{L}_{KL}$ term regularizes the posterior toward the prior; $\beta=1$ gives the standard ELBO objective. We did not explore alternative KL-weight schedules in the reported experiments.

\subsection{Temporal Position Encoding}
\label{app:position}
For sequence position $pos$ and embedding coordinate $i$, we use the small-embedding-friendly sinusoidal variant
\begin{equation}
PE_{(pos,2i)}=\sin(pos\,\omega_i^{\mathrm{new}}),\qquad
PE_{(pos,2i+1)}=\cos(pos\,\omega_i^{\mathrm{new}}),
\end{equation}
where $\omega_i^{\mathrm{new}}=\omega_i d/L$ and $\omega_i=10000^{-2i/d}$. The scaling adapts the standard sinusoid to the short interaction vocabulary and sequence length used in our setting.

\section*{Ethical Considerations}
We identify no ethical implications specific to the proposed optimization method beyond the standard privacy, fairness, and experimentation considerations of large-scale recommendation systems; deployments should follow applicable data-governance and review processes.

\bibliographystyle{ACM-Reference-Format}
\bibliography{reference}

\end{document}